\pdfoutput=1

\documentclass{article}
\usepackage{ijcai19}

\usepackage{times}
\usepackage{soul}
\usepackage{url}
\usepackage[utf8]{inputenc}
\usepackage[small]{caption}
\usepackage{graphicx}
\usepackage{amsmath}
\usepackage{booktabs}
\usepackage{algorithm}
\usepackage{algorithmic}
\usepackage{multirow}
\usepackage{mathtools}

\usepackage{xcolor}

\newcommand{\exF}{\mathbf{F}_{ex}}
\newcommand{\fsF}{\mathbf{F}_{fs}}
\newcommand{\pD}{\mathbf{D}}
\newcommand{\rD}{\mathbf{D}_{res}}

\title{Structure-Aware Residual Pyramid Network for Monocular Depth Estimation}

\author{
Xiaotian Chen
\and
Xuejin Chen\footnote{Corresponding author}\And
Zheng-Jun Zha
\affiliations
National Engineering Laboratory for Brain-inspired Intelligence Technology and Application\\
University of Science and Technology of China\\
\emails
ustcxt@mail.ustc.edu.cn,
\{xjchen99, zhazj\}@ustc.edu.cn
}

\begin{document}
	\maketitle
	\begin{abstract}
		Monocular depth estimation is an essential task for scene understanding. 
		The underlying structure of objects and stuff in a complex scene is critical to recovering accurate and visually-pleasing depth maps.
		Global structure conveys scene layouts, while local structure reflects shape details. 
		Recently developed approaches based on convolutional neural networks (CNNs) significantly improve the performance of depth estimation. 
		However, few of them take into account multi-scale structures in complex scenes.
		In this paper, we propose a Structure-Aware Residual Pyramid Network (SARPN) to exploit multi-scale structures for accurate depth prediction. 
		We propose a Residual Pyramid Decoder (RPD) which expresses global scene structure in upper levels to represent layouts, and local structure in lower levels to present shape details. 
		At each level, we propose Residual Refinement Modules (RRM) that predict residual maps to progressively add finer structures on the coarser structure predicted at the upper level.
		In order to fully exploit multi-scale image features, an Adaptive Dense Feature Fusion (ADFF) module, which adaptively fuses effective features from all scales for inferring structures of each scale, is introduced. 
		Experiment results on the challenging NYU-Depth v2 dataset demonstrate that our proposed approach achieves state-of-the-art performance in both qualitative and quantitative evaluation. The code is available at \url{https://github.com/Xt-Chen/SARPN}.
	\end{abstract}	
	\section{Introduction}
\label{sec:intro}

\begin{figure}[htp]
	\centering 
	\includegraphics[width = 0.98\columnwidth]{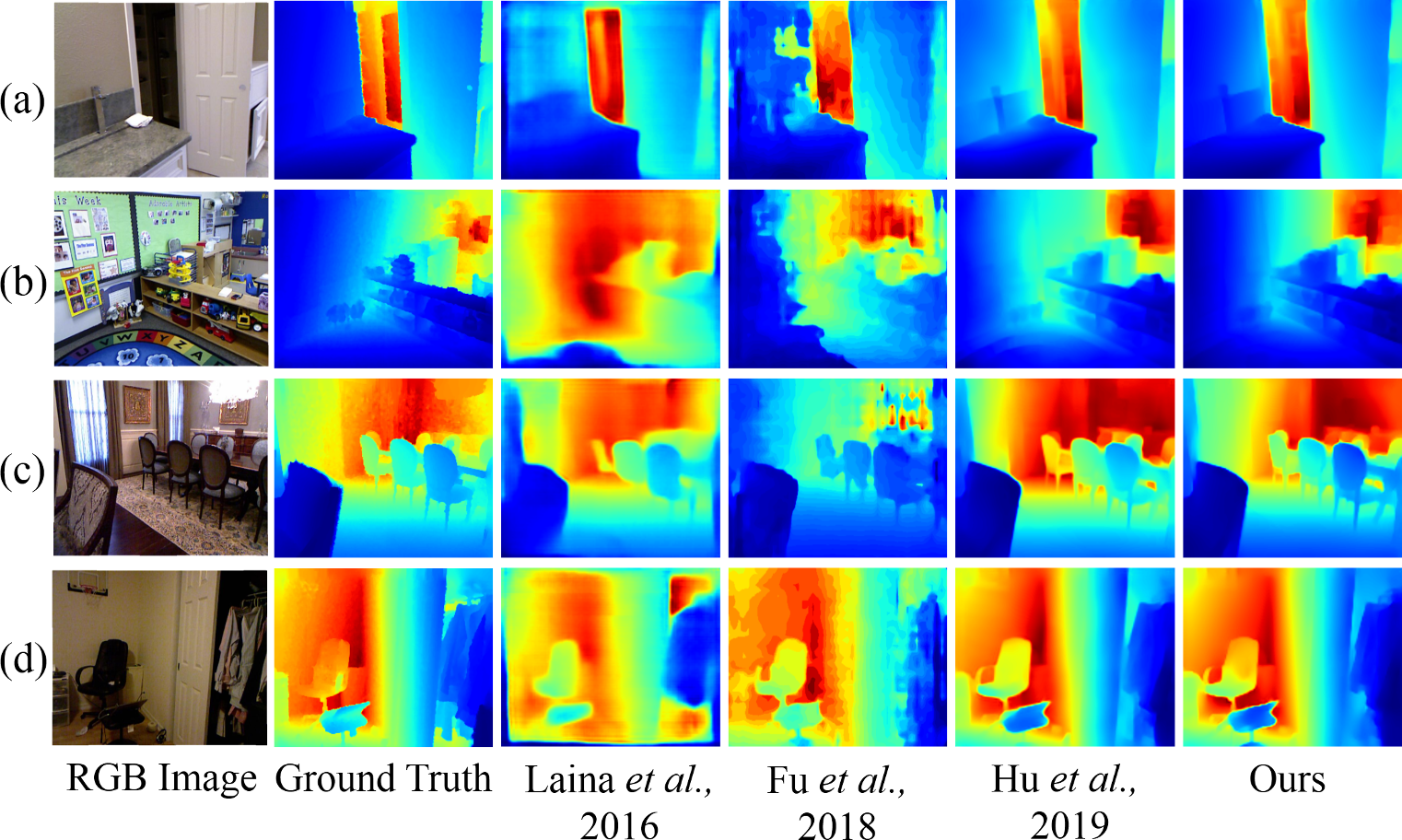}
	\caption{Problems in depth prediction: (a)(b) inaccurate depth values on large planar regions, such as walls. (c)(d) blurry boundaries and missing details (chair legs). Our approach simultaneously recovers large planar structures and object details.}
	\label{fig:intro-compare}
\end{figure}

Monocular depth estimation, which aims to predict the depth value of each pixel from a given RGB image, is crucial for understanding scene geometry, and can be applied to facilitate other vision tasks, such as semantic segmentation~\cite{park2017rdfnet} and hand tracking~\cite{qian2014realtime}.
It is an ill-posed problem because of the inherent ambiguity due to perspective projection.
Recently, CNN-based approaches have achieved significant success in monocular depth estimation \cite{laina2016deeper,fu2018deep,xu2018structured,hao2018detail,hu2019revisiting}. 
To resolve the ambiguity, they typically employ an encoder-decoder architecture to implicitly fuse features that represents object appearance, geometry, semantics, spatial relations, etc.  
The encoder gradually extracts multi-scale features, and the decoder employs multi-stage upsampling as well as shortcut connections to restore object details in high-resolution predictions. 

Though a great improvement on average pixel-wise metrics has been made, the underlying structure of objects and stuff is not well preserved by current CNN-based methods. 
The problem becomes especially challenging when the size of objects and stuff varies widely in complex scenes.
As Figure \ref{fig:intro-compare} shows, it is challenging for existing approaches to accurately recover the large-scale geometry (walls) and local details (boundaries and small parts) at the same time.
This inaccurate inference at regions of diverse scales motivates us to fully exploit the hierarchical scene structure in depth prediction.
Scene structure, depicting the organization and arrangement of multiple interrelated elements in a complex scene, varies widely according to the element type. 
The global structure represents the spatial arrangement of large-size elements such as walls, floors, and furniture objects. 
Local structure describes geometric details of objects and their parts.
The natural hierarchy of scene structure provides essential constraints between the depth values of pixels in multiple scales. 
Although previous CNN-based techniques extract multi-scale image features and gradually fuse them to predict a depth map, the underlying hierarchical structure of the scene has not been taken into account. 

In this paper, we introduce a Structure-Aware Residual Pyramid Network (SARPN) to fully exploit scene structures in multiple scales for depth prediction. A Residual Pyramid Decoder (RPD) is proposed to predict multi-scale depth maps in a coarse-to-fine manner.
Depth maps in upper levels in the pyramid represent the global scene structure, while depth maps in lower levels capture more local structures of objects or parts. To convey the global structure and constrain the generation of finer details, we proposed a residual refinement module to predict residual depth maps, which progressively add details on the scene structure on a larger scale. 
In order to fuse multi-scale features extracted from the input image for residual prediction, we propose an Adaptive Dense Feature Fusion (ADFF) module to adaptively select more effective features for each scale. 
Integrating the residual pyramid decoder and adaptive dense feature fusion module, our method 
simultaneously preserves the hierarchical scene structures and produces accurate depth estimation for both large-size shapes and fine details of small object parts, as Figure \ref{fig:intro-compare} shows.
Our contributions are summarized as follows:
\begin{itemize}
	\item We propose a Structure-Aware Residual Pyramid Network (SARPN), which takes the underlying scene structure in multiple scales into account for accurate depth prediction.
	\item Our Adaptive Dense Feature Fusion (ADFF) module adaptively selects features from all scales to predict residual depths at different structure scales.
	\item The proposed method achieves state-of-the-art performance on the challenging NYUD v2 dataset. More importantly, the visual quality of recovered depth maps is significantly improved.
\end{itemize}
	\section{Related Work}
In recent years, CNNs have become the most successful techniques for various visual tasks, and were firstly used for monocular depth estimation~\cite{eigen2014prediction} in a multiple scale scheme.
Later on, fully convolutional network (FCN) was proposed for semantic segmentation~\cite{long2015fully} and has been widely used in many dense prediction tasks, including depth estimation. 

When FCN-based architecture was first adopted for depth estimation, the resolution and accuracy ware largely improved by using ResNet to extract features and up-projection blocks~\cite{laina2016deeper}.
In order to improve the quality of depth estimation for local details, many strategies have been introduced. 
Applying conditional random field as post-processing~\cite{li2015depth} or integrating it in CNNs~\cite{xu2017multi} largely improves the prediction quality for small objects.
Later, an attention model is integrated to improve the estimation performance~\cite{xu2018structured}. 
Multi-scale architecture becomes a common solution to avoid the loss of local details caused by spatial pooling and convolutions \cite{fu2018deep}.
Instead of multi-scale network structure, dilated 
convolution is used to extract multi-scale features for depth estimation \cite{hao2018detail}.
Hu et al. \shortcite{hu2019revisiting} proposed an effective multi-scale feature fusion module to produce clear object boundaries.
Although these methods have achieved remarkable results by fusing multi-scale features, they still face the problem of inaccurate prediction for complex scenes of which the structure varies largely in scales, from large room layout to fine object details.

In order to better restore structure details, a few methods design new loss functions to explicitly constrain scene geometry. 
Zheng et al.~\shortcite{zheng2018net} proposed an order-sensitive softmax loss to constrain global layouts. Similarly, Fu et al.~\shortcite{fu2018deep} used an ordinary regression loss. 
With respect to clear boundaries and details, a loss function is designed by combining depth, surface normal and gradient in a local neighborhood of depth maps~\cite{hu2019revisiting}. 

Due to the strong correlation between many visual tasks, such as depth estimation, semantic segmentation, and normal estimation, many approaches employ a joint task learning framework. A multi-scale CNN was designed to simultaneously perform semantic segmentation, depth estimation, and normal estimation \cite{eigen2015predicting}. 
A set of intermediate auxiliary tasks are utilized to guide the final depth estimation and semantic segmentation \cite{Xu_2018_CVPR}. 
Zhang et al. proposed a novel joint task-recursive learning method to recursively refine the results of depth estimation and semantic segmentation~\cite{Zhang_2018_ECCV}. 
A synergy network is proposed to automatically learn information sharing strategy between depth estimation and semantic segmentation~\cite{jiao2018look}.
Moreover, based on the observed long-tail distribution of depth values, an attention-driven loss is also designed to improve the accuracy~\cite{jiao2018look}.

	\section{Methodology}
\label{sec:method}

\begin{figure*}[ht]
	\centering 
	\includegraphics[width=\textwidth]{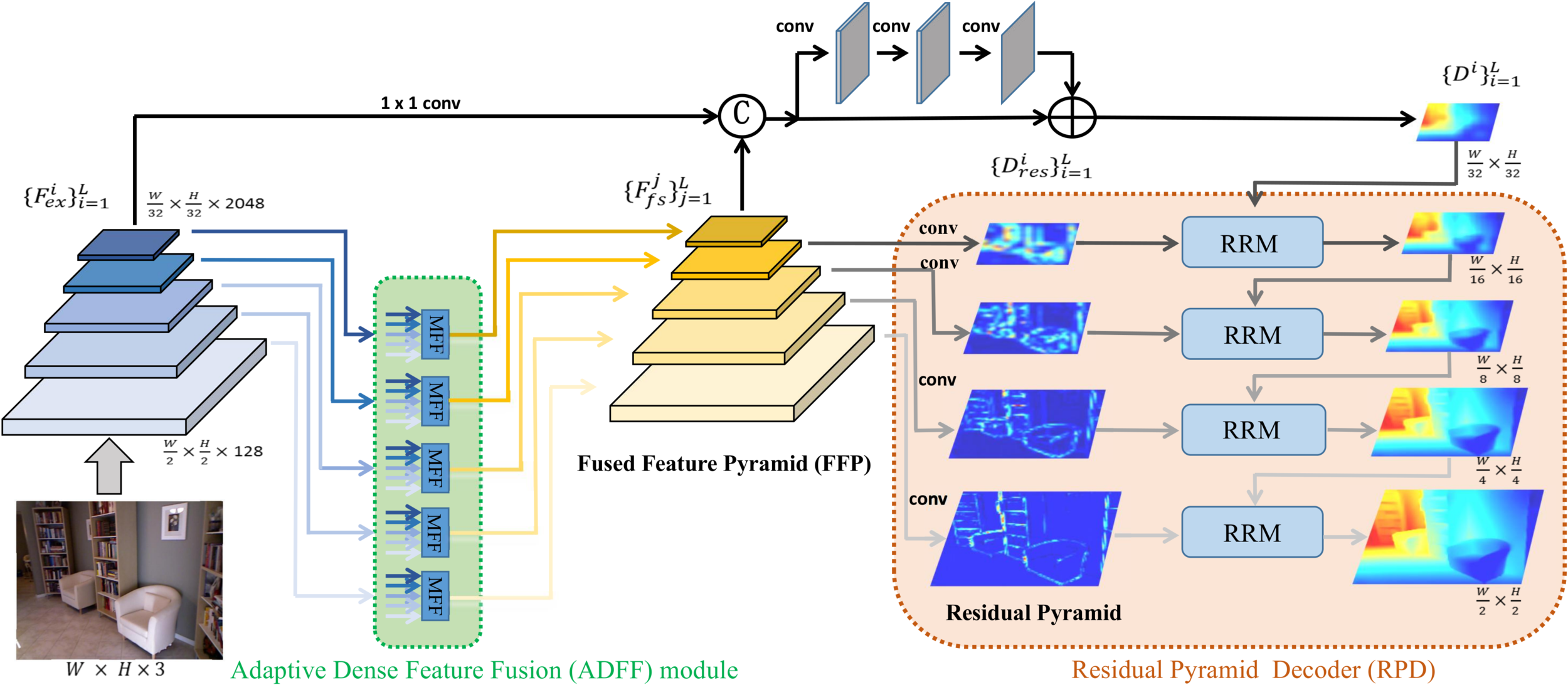}
	\caption{The network architecture. Our Structure-Aware Residual Pyramid Network consists of an encoder which extracts multi-scale visual features, a Residual Pyramid Decoder (RPD) which progressively infers depth maps in a coarse-to-fine manner, and an Adaptive Dense Feature Fusion (ADFF) module for dense feature fusion. The residual pyramid effectively adds structure details in each level based on the scene layout predicted at a coarser level. }
	\label{fig:NetworkArchitecture}
\end{figure*}

Our network consists of three main parts: an encoder for multi-scale feature extraction, an adaptive dense feature fusion module, and a residual pyramid decoder, as Figure~\ref{fig:NetworkArchitecture} shows.
We first introduce the network architecture in Sec.~\ref{sec:network}. The residual pyramid decoder and adaptive dense feature fusion module are explained in Sec.~\ref{sec:RPD} and \ref{sec:ADFF}, respectively. 

\subsection{Structure-Aware Residual Pyramid Network}
\label{sec:network}

Our approach begins with an encoder which extracts multi-scale features $\{\exF^{i}\}_{i=1}^L$ from the input image, where $\exF^{i}$ indicates the feature maps extracted at the $i$-th level. $L$ is the number of layers in our network. Following the state-of-the-art approach~\cite{hu2019revisiting}, we use SENet~\cite{hu2018squeeze} as the backbone of our encoder. It extracts more effective features by re-weighting features of different channels. Given an input image with size $W \times H$, the size of these feature maps are respectively $[\frac{W}{2^i},\frac{H}{2^i}]$, and they carry both high-level semantic information and low-level detail information. 
Then, these multi-scale feature maps are simultaneously fed to our dense feature fusion module to produce a Fused Feature Pyramid~(FFP). 
These feature maps in FFP are represented by $\{\fsF^i\}_{i=1}^L$, where $\fsF^i$ indicates the fused feature maps at the $i$-{th} level of the pyramid of fused features. 

In the decoder part, different from the previous methods that directly predict a depth map by sequentially upsampling feature maps~\cite{laina2016deeper,hu2019revisiting}, our residual pyramid progressively predicts multiple depth maps in a coarse-to-fine manner. 
The depth map at the top level with size $\frac{W}{32} \times \frac{H}{32}$ is predicted first as the initial scene layout. 
We utilize a $1\times1$ convolution operation to reduce the channel number of the feature maps $\exF^L$ to the same as the channel number of feature maps $\fsF^L$ of fused feature pyramid and concatenate them together. 
A residual block is used to predict a depth map $\pD^L$ in size of $[\frac{W}{2^L}, \frac{H}{2^L}]$ from the concatenated feature maps. 
Then we gradually refine the depth prediction by our proposed residual pyramid decoder.

\subsection{Residual Pyramid Decoder}
\label{sec:RPD}
\begin{figure}[t]
	\centering 
	\includegraphics[width = 0.9\columnwidth]{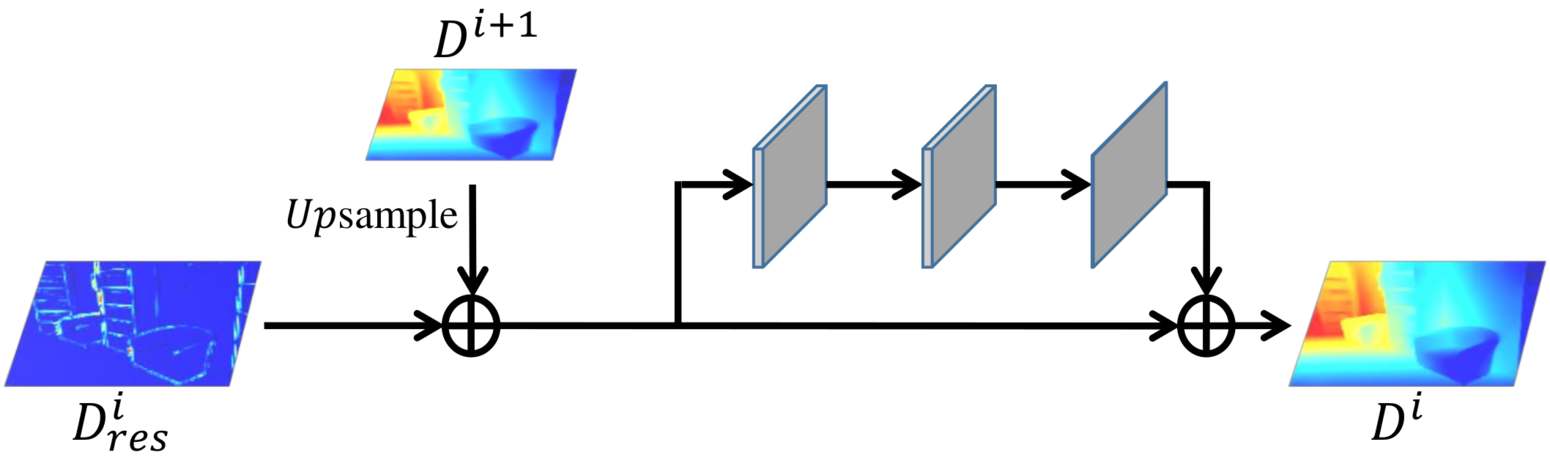}
	\caption{A Residual Refinement Module (RRM) for the $i$-{th} level.}
	\label{fig:residualRM}
\end{figure}

Our residual pyramid decoder predicts depth maps of multiple scales in order to restore the hierarchical scene structures in a coarse-to-fine manner. 
As shown in Figure~\ref{fig:NetworkArchitecture}, the depth maps in lower resolutions depicts more global scene layout, while the depth maps in higher resolutions contain more structure details.
In each level of the pyramid decoder, we predict a residual map instead of a dense depth map from fused image features in FFP.
The residual map and the depth map predicted at the upper level are integrated together to produce a refined depth map in the current scale using our Residual Refinement Module (RRM). 
The components of each RRM are shown in Figure \ref{fig:residualRM}.
The depth map $\pD^{i+1}$ predicted at the upper scale is upsampled to the current scale by bilinear interpolation. 
A residual depth map $\rD^{i}$ is generated by utilizing the fused features $\fsF^{i}$. 
After adding the residual map and the upsampled depth map, a residual block, which contains three convolutional layers, is employed to refine the prediction and outputs a depth map $\pD^{i}$ at the $i$-th scale.
This residual architecture induces our network to effectively represent the structure details at each scale and hierarchically refine scene structures.
Meanwhile, the global scene layout is well preserved by our residual pyramid decoder.

\begin{table*}[htb]
\centering
\begin{tabular}{l|c|c|c|c|c|c}
\toprule
Method  & REL & RMS & $\log{10}$ & $\delta < 1.25$ & $\delta < 1.25^2$ & $\delta < 1.25^3$ \\
\midrule
Ladicky et al. \shortcite{ladicky2014pulling}     & -  & -  & -  & 0.542 & 0.829  & 0.941     \\
\hline
Li et al. \shortcite{li2015depth}                 & 0.232  & 0.821  & 0.094  & 0.621 & 0.886  & 0.968     \\
\hline
Eigen et al. \shortcite{eigen2014prediction}      & 0.215  & 0.907  & -  & 0.611 & 0.887  & 0.971     \\
\hline
Laina et al. \shortcite{laina2016deeper}          & 0.127  & 0.573 & 0.055  & 0.811 & 0.953  & 0.988     \\
\hline
Xu et al. \shortcite{xu2017multi}                 & 0.121  & 0.586  & 0.052  & 0.811 & 0.954  & 0.987   \\
\hline
Xu et al. \shortcite{xu2018structured}            & 0.125  & 0.593  & 0.057  & 0.806 &  0.952  &  0.986      \\
\hline
Hao et al. \shortcite{hao2018detail}              & 0.127  & 0.555  & 0.053  & 0.841 & 0.966  & 0.991     \\
\hline
Fu et al. \shortcite{fu2018deep}                  & 0.115  & \emph{0.509}  & 0.051  & 0.828 & 0.965  & 0.992    \\
\hline
Qi et al. \shortcite{qi2018geonet}                & 0.128  & 0.569  & 0.057  & 0.834 & 0.960  & 0.990    \\
\hline
Jiao et al. \shortcite{jiao2018look}              & 0.126  & \textbf{0.416}  & 0.050  & 0.868 & 0.973  & 0.993     \\
\hline
Hu et al. \shortcite{hu2019revisiting}            & 0.115  & 0.530  & 0.050  & 0.866 & 0.975  & 0.993     \\
\hline
\hline 
Our Baseline                                     & 0.123  & 0.547   & 0.052  & 0.854 & 0.969  & 0.992     \\
\hline
Our Baseline + RPD                               & 0.115  & 0.528  & 0.050  & 0.871 & 0.975  & 0.993     \\
\hline
\textbf{Ours}: Baseline + RPD + ADFF                                & \textbf{0.111}  & 0.514  & \textbf{0.048}  & \textbf{0.878} & \textbf{0.977}  & \textbf{0.994}     \\
\hline
\hline
Eigen and Fergus \shortcite{eigen2015predicting}*     & 0.158  & 0.641  & -  & 0.769 & 0.950  &  0.988    \\
Xu et al. \shortcite{Xu_2018_CVPR}*                   & 0.120  & 0.582  & 0.055 & 0.817 & 0.954 & 0.987   \\
Zhang et al. \shortcite{Zhang_2018_ECCV}*             & 0.144  & \emph{0.501}  & -     & 0.815 & 0.962 & 0.992   \\ 
Jiao et al. \shortcite{jiao2018look}*                  & \emph{0.098}  & \emph{0.329}  & \emph{0.040} & \emph{0.917} & \emph{0.983} & \emph{0.996}   \\ 
\bottomrule
\end{tabular}
\caption{Comparisons with state-of-the-art depth estimation approaches on NYUD v2 Dataset. Note that joint task learning is employed in the methods marked by *. The best results on each metric among the single-task approaches are marked in bold type. The results better than ours are marked in italics.}
\label{tab:comparison}
\end{table*}

\subsection{Adaptive Dense Feature Fusion}
\label{sec:ADFF}

In general, due to pooling operations and convolution operations with strides in CNNs, a large amount of low-level visual features are lost. 
As a result, it is difficult for the decoder to recover the lost low-level structure details. 
However, both low-level features and high-level features are critical for predicting residual maps in all layers, because the residual maps convey additional details on a global scene structure, as the residual pyramid illustrates in Figure~\ref{fig:NetworkArchitecture}. 
In order to provide sufficient information for the prediction of a residual map in each level, we propose an Adaptive Dense Feature Fusion (ADFF) module.
This dense fusion module consists of $L$ Multi-scale Feature Fusion (MFF) modules to predict $L$ fused feature maps, which compose a fused feature pyramid for residual prediction.  

In each layer, the MFF adaptively selects eligible features from all feature scales when predicting the depth map for each individual scale. 
We follow the detailed implementation of MFF proposed in~\cite{hu2019revisiting}. 
The $L$ feature maps $\{\exF^{i}\}_{i=1,\ldots,L}$ are first resized to the resolution of current scale using bilinear interpolation and refined with a residual refine block.
The refined feature maps are concatenated and fed into a conv-layer to reduce the number of channels. 

\subsection{Loss Function}
\label{sec:implementation}

In order to train our residual pyramid network for predicting accurate depth maps while preserving scene structures in various scales, we compute the difference between the predicted depth map $\pD^{i}$ and the ground-truth $\mathbf{G}^i$ at each scale and combine the losses of all scales together. 
For each scale, we follow the definition of the loss function proposed in \cite{hu2019revisiting}. 
It consists of three terms, $l_{depth}$ considering the pixel-wise difference between the predicted depth $\pD^{l}$ and the ground truth $G^{l}$, $l_{grad}$ which penalizes errors round edges, and $l_{normal}$ to further improve fine details. 
Combing all the $L$ scales, our loss function for the entire network is formulated as 
\begin{equation}
L = \sum_{i=1}^L {l_{depth}^i + l_{grad}^i +  l_{normal}^i}.
\end{equation}

	\section{Experiments}
\label{sec:results}

To demonstrate the effectiveness of the proposed approach, we evaluate our approach on the challenging NYUD v2 dataset~\cite{silberman2012indoor}. We compare our approach with a couple of state-of-the-art approaches and show the superiority of the proposed method on both quantitative and qualitative evaluations.

\begin{figure}[ht]
	\centering
	\includegraphics[width=\columnwidth]{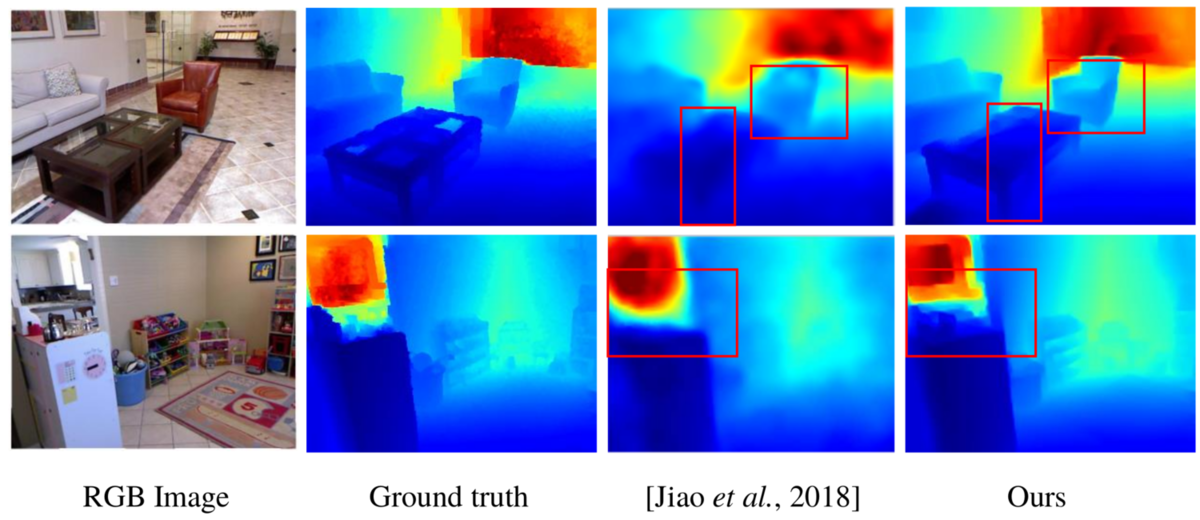}
	\caption{Comparison with [Jiao et al., 2018]. The depth maps predicted by our method preserve much more accurate depth around object boundaries and keep finer structures, as highlighted in the boxes. }
	\label{fig:compare-jiao}
\end{figure}

\subsection{Experimental Setup}
The NYU-Depth v2 dataset~\cite{silberman2012indoor} contains 464 video sequences of indoor scenes captured with Microsoft Kinect. 
654 aligned RGB-Depth pairs are provided for testing depth estimation methods for indoor scenes. 
All images have a resolution of 640 $\times$ 480. 
To training our network, we use the training dataset which contains 50K RGBD images, select and then augment in the same way as \cite{hu2019revisiting}. 
Each image is downsampled to $320 \times 240$ using bilinear interpolation, and then center-cropped to 304 $\times$ 228.
The predicted depth maps are in a resolution of $152\times 114$.
For testing, the predicted depth maps are upsampled to match the size of the corresponding ground truth using bilinear interpolation.
\begin{figure*}  
	\centering 
	\includegraphics[width =0.87\textwidth]{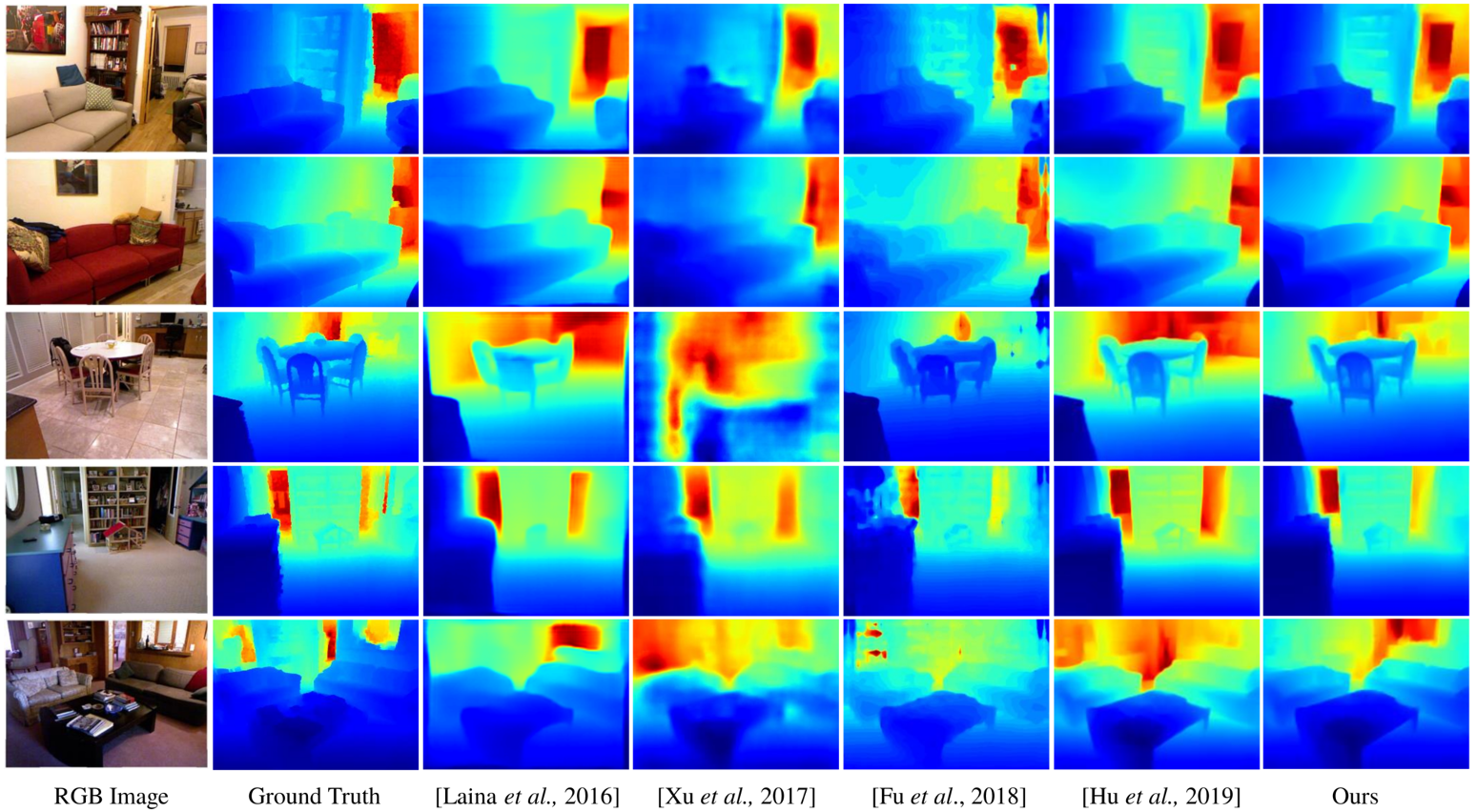}
	\caption{Qualitative results on the NYUD2 dataset.}
	\label{fig:visual-comp}
\end{figure*}

We implement the proposed model using PyTorch \cite{paszke2017automatic}.
The encoder, SENet, is initialized by a model pretrained on ImageNet~\cite{deng2009imagenet}. 
The other layers in our network are randomly initialized. 
We use a step learning rate decay policy with Adam optimizer, and starting from an initial learning rate of $l_{init}=10^{-4}$. It is reduced to $10\%$ every 5 epochs. 
We use $\beta _1=0.9$, $\beta_2=0.999$, and weight decay as $10^{-4}$. 
The proposed network was trained for 20 epochs with a batch size of 6.

\subsection{Performance Comparison} 
\subsubsection{Quantitative Evaluation} 
Following previous studies, we adopt four metrics including average relative error (REL), root mean squared error (RMS), mean $\log 10$ error (log 10), and accuracy with three thresholds, to quantitatively evaluate our depth estimation performance.
Table~\ref{tab:comparison} shows the results of our SARPN and recent approaches. 
Among the approaches of single task learning, our approach performs the best on REL, log 10 error, and accuracy with three thresholds.
We are in the third position with respect to RMS.
We speculate that the methods~\cite{fu2018deep,jiao2018look} pay more attention to the absolute pixel-wise accuracy when designing their networks and loss functions, ignoring fine structures of target scenes. 
As a result, these methods achieve higher performance in RMS, but performs worse on the REL metric and other metrics. 

We also compare our method with four approaches that employ joint task learning~\cite{eigen2015predicting,Xu_2018_CVPR,Zhang_2018_ECCV,jiao2018look}.
The results demonstrated that our method outperforms three methods and achieves comparative performance with \cite{jiao2018look}, even they use a large number of extra labels for semantic segmentation during the training process.
Moreover, the depth maps produced by \cite{jiao2018look} present very blurry object boundaries and miss geometric details. 
We compare the predicted depth maps in Figure~\ref{fig:compare-jiao} to demonstrate the capability of our method on restoring clear object boundaries and finer details.

\begin{table}[tb]
	\centering
	\begin{tabular}{|c|c|c|c|c|}
		\hline
		Thres  & Method & Prec & Recall & F1  \\
		\hline
		\multirow{5}*{$0.25$}                          & \cite{laina2016deeper}   & 0.489   & 0.435  & 0.454      \\
		~ &\cite{Xu_2018_CVPR}    & 0.516  &  0.400  & 0.436      \\
		~ & \cite{fu2018deep}     & 0.320  & \textbf{0.583}  & 0.402       \\
		~ & \cite{hu2019revisiting} & 0.644 & 0.508 & 0.562 \\
		~ & Ours  & \textbf{0.645} & 0.520 & \textbf{0.570}\\
		\hline                          
		\multirow{5}*{$  0.5$}                          &\cite{laina2016deeper}   & 0.536   & 0.422  & 0.463      \\
		~ &\cite{Xu_2018_CVPR}    & 0.600  &  0.366  & 0.439      \\
		~ &\cite{fu2018deep}     & 0.316  &0.473    &0.412      \\
		~ &\cite{hu2019revisiting} & \textbf{0.668} & 0.505 & 0.568 \\
		~ &Ours  & 0.663 & \textbf{0.523} & \textbf{0.578}\\
		\hline                          
		\multirow{5}*{$ 1.0$}                          &\cite{laina2016deeper} & 0.670   & 0.479  & 0.548      \\
		~ &\cite{Xu_2018_CVPR}    & \textbf{0.794}  &  0.407  & 0.525      \\
		~ &\cite{fu2018deep}   & 0.483  &0.512    &0.485      \\
		~ &\cite{hu2019revisiting} &0.759 &0.540 &0.623 \\
		~ &Ours  &0.749& \textbf{0.554} & \textbf{0.630}\\
		\hline 
	\end{tabular}
	\caption{Accuracy of recovered edge pixels in depth maps under different thresholds.}
	\label{tab:edgeaccuracy}
\end{table}
We also analyze the contribution of each component in our proposed network. 
We use a simple UNet-like architecture as our baseline, where SENet \cite{hu2018squeeze} is employed as the backbone of our encoder. 
The decoder in our baseline employs a multi-stage upsampling scheme to recover a depth map. 
A variant (baseline+RPD) is implemented by adding the proposed RPD on the baseline model.
As shown in Table \ref{tab:comparison}, the performance is gradually improved by incorporating RPD and ADFF. More specifically, after adding the proposed RPD, performance among all the metrics are improved by a large margin from the baseline, while REL decreases by $6.5\%$, RMS decreases by $3.5\%$, $\log 10$ error decreases by $3.8\%$. 
After adding the ADFF module, the performance is further improved, while REL decreases by $3.5\%$, RMS decreases by $2.7\%$ and $\log 10$ error decreases by $4\%$.  

In order to prove the effectiveness of our method on preserving object details, we also compute edge accuracy to measure the quality of recovered edge details, same as~\cite{hu2019revisiting}. Precision, Recall, and F1 score are computed according to edge pixels in the ground truth map. From Table~\ref{tab:edgeaccuracy}, we can see that our F1 score surpasses all other methods under three different thresholds. This indicates that our method restores the most structure details.
\subsubsection{Qualitative Evaluation}
We compare a series of depth maps predicted by our method and other state-of-the-art methods~\cite{laina2016deeper,xu2017multi,fu2018deep,hu2019revisiting} in Figure \ref{fig:visual-comp}.
It can be seen that the depth maps predicted by our method are visually better than other methods.
Scene structures are well preserved in different scales, especially for large planar regions and object details.
For example, our method predicts accurate geometric details for the bookshelf in the first row, the chair in the third row, and the sofa in the fifth row. 
For large planar regions (the upper-left wall in the second row, and the floor of the third), our method also generates better results. 

\begin{figure} 
	\centering 
	\includegraphics[scale=0.23]{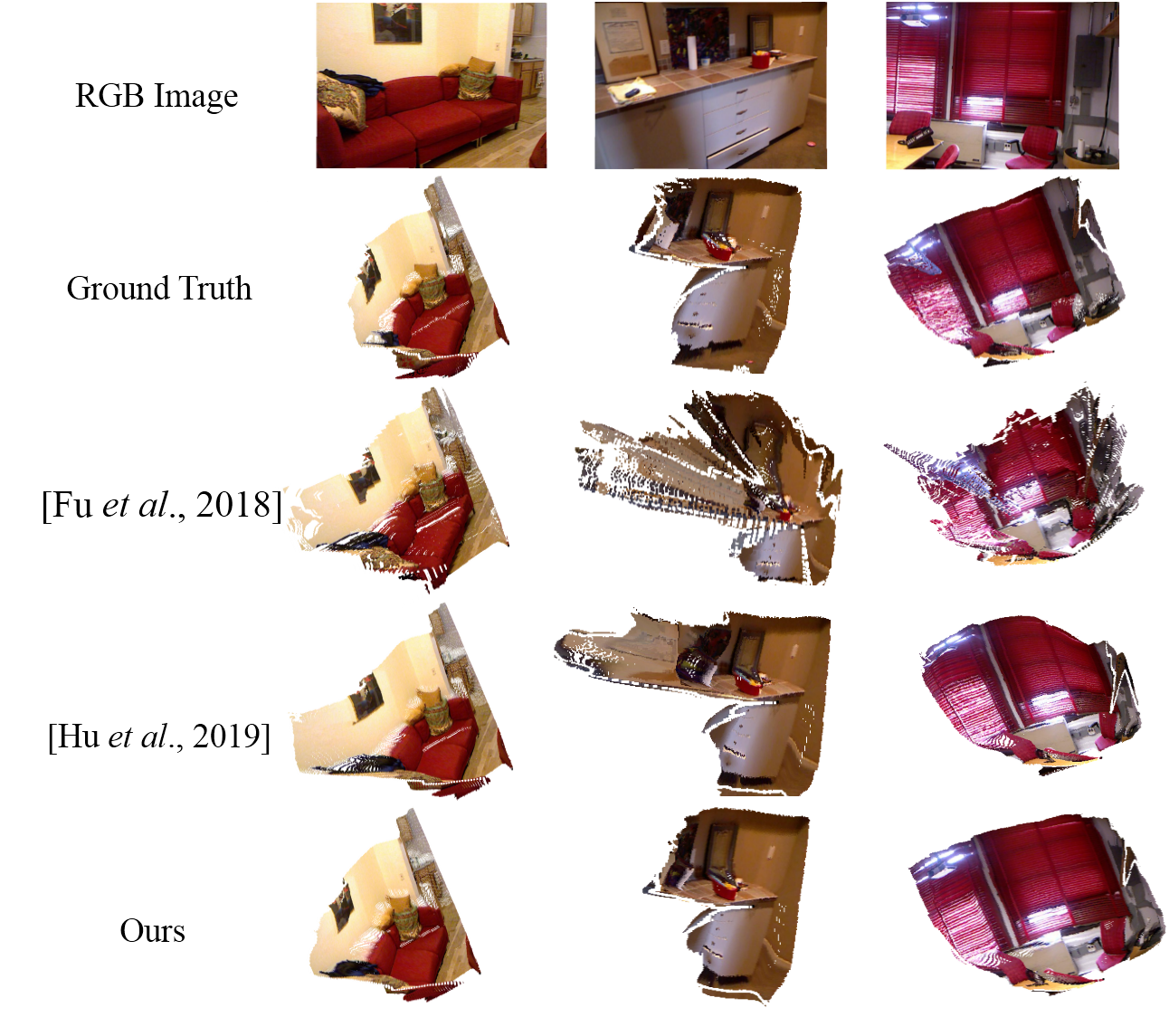}
	\caption{3D projection from predicted depth maps. Our method better preserves the scene structure of various scales, especially the flat shape of large planar regions.}
	\label{fig:Rendering}
\end{figure}

To better illustrate the capability of our method on preserving scene structure of large planar regions, we project the predicted depth maps as 3D point clouds and render them in novel views.
As Figure~\ref{fig:Rendering} shows, our reprojected results are the closest to ground truth. In particular, the large wall regions recovered by our method are much more flat, while other methods suffer from severe distortions. 

\begin{figure} 
	\centering 
	\includegraphics[scale=1.1]{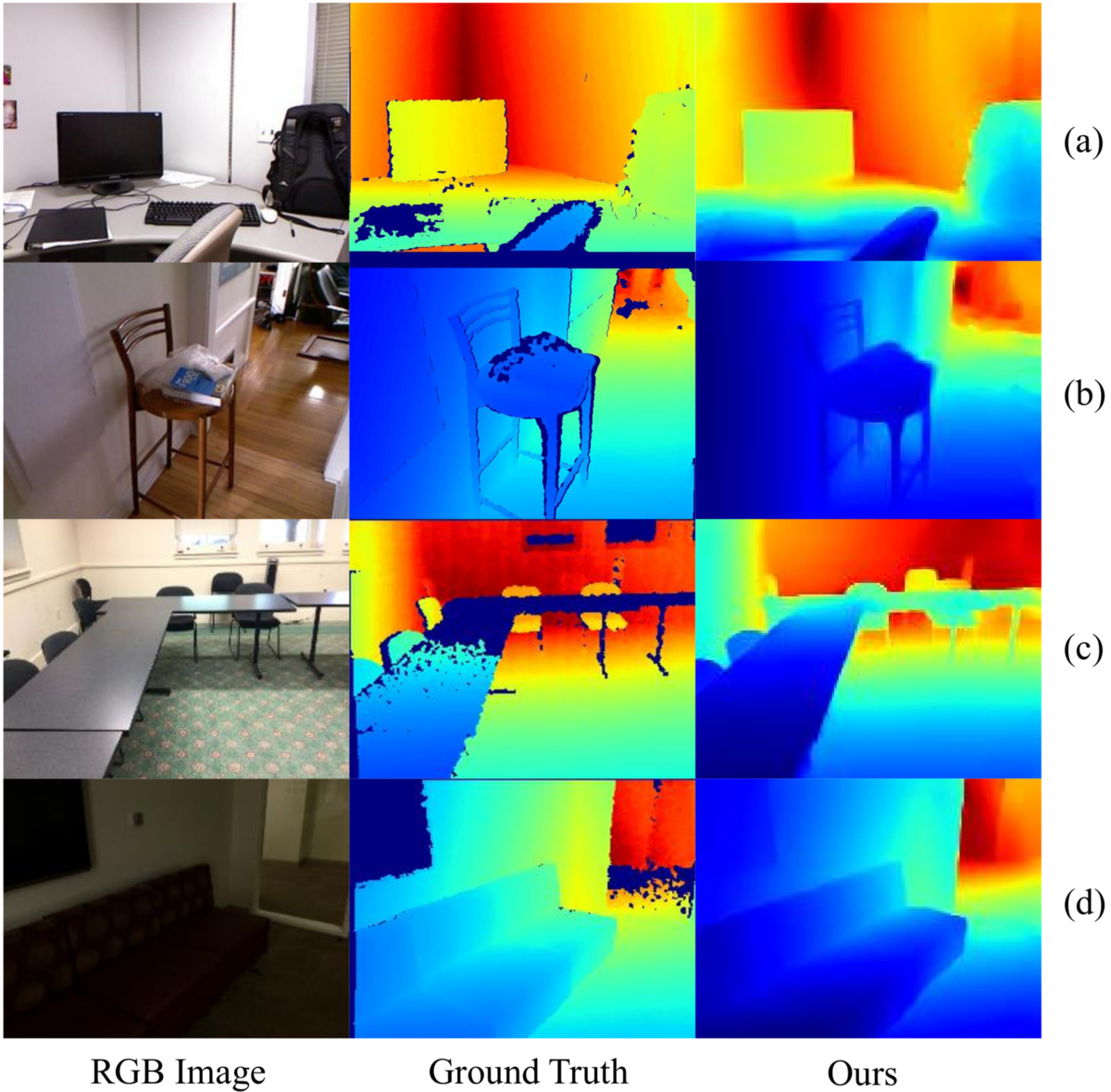}
	\caption{More results by applying our model on SUN-RGBD dataset (a)(b) and ScanNet dataset (c)(d).}
	\label{fig:newdataset}
\end{figure}

\subsubsection{Model Generalization} 
In addition to the NYUD v2 dataset, we further explore the generalization ability of our proposed network on other datasets.
We test our network, which is trained on the NYUD v2 dataset only, on ScanNet dataset~\cite{dai2017scannet} and SUN-RGBD dataset~\cite{song2015sun}, which contain more diverse RGBD data.  
As shown in Figure~\ref{fig:newdataset}, even the data distribution of these two datasets and NYU Depth v2 is greatly different, our method could recover structures in various scales, including smooth large planar regions and object details. 
Moreover, our method also fills holes in the ground truth depth map automatically while maintains the scene structure.

	\section{Conclusion}
	
	In this paper, we propose a Structure-Aware Residual Pyramid Network for accurate monocular depth estimation. 
	A residual pyramid decoder is introduced to predict multi-scale depth maps, which takes the underlying hierarchical scene structures into account.
	The residual pyramid induces our network to progressively add finer structures at a specific scale while preserving the coarser layout predicted at the upper level. 
	Meanwhile, by using the proposed adaptive dense feature fusion module, the image features from all scales are adaptively fused when predicting the residual depth map for each scale. Experiment results demonstrate that our method achieves state-of-the-art performance in both quantitative and qualitative evaluation. 

	\section*{Acknowledgements}
	This work was supported by the National Key Research \& Development Plan of China under Grant 2018YFC0307905, the National Natural Science Foundation of China (NSFC) under Grants 61632006, 61622211, and 61620106009, the Priority Research Program of Chinese Academy of Sciences under Grant XDB06040900, as well as the Fundamental Research Funds for the Central Universities under Grant WK3490000003 and WK2100100030.

	\bibliographystyle{named}
	\bibliography{depth-ijcai19}
	
\end{document}